\documentclass[12pt,twoside]{article}
\usepackage{latexsym}
\def\,{\mskip 3mu} \def\>{\mskip 4mu plus 2mu minus 4mu} \def\;{\mskip 5mu plus 5mu} \def\!{\mskip-3mu}
\def\dispmuskip{\thinmuskip= 3mu plus 0mu minus 2mu \medmuskip=  4mu plus 2mu minus 2mu \thickmuskip=5mu plus 5mu minus 2mu}
\def\textmuskip{\thinmuskip= 0mu                    \medmuskip=  1mu plus 1mu minus 1mu \thickmuskip=2mu plus 3mu minus 1mu}
\textmuskip
\def\beq{\dispmuskip\begin{equation}}    \def\eeq{\end{equation}\textmuskip}
\def\beqn{\dispmuskip\begin{displaymath}}\def\eeqn{\end{displaymath}\textmuskip}
\def\bqa{\dispmuskip\begin{eqnarray}}    \def\eqa{\end{eqnarray}\textmuskip}
\def\bqan{\dispmuskip\begin{eqnarray*}}  \def\eqan{\end{eqnarray*}\textmuskip}

\newtheorem{theorem}{Theorem}
\newtheorem{corollary}[theorem]{Corollary}
\newtheorem{lemma}[theorem]{Lemma}

\newenvironment{keywords}{\centerline{\bf\small
Keywords}\begin{quote}\small}{\par\end{quote}\vskip 1ex}
\def\paradot#1{\vspace{1ex plus 1ex minus 0.5ex}\noindent{\bf{#1.}}}
\def\paranodot#1{\vspace{1ex plus 1ex minus 0.5ex}\noindent{\bf{#1}}}
\def\eps{\varepsilon}
\def\nq{\hspace{-1em}}
\def\qed{\hspace*{\fill}$\Box\quad$\\}
\def\odt{{\textstyle{1\over 2}}}
\def\SetR{I\!\!R}
\def\SetN{I\!\!N}

\def\D{{\cal D}}
\def\S{{\cal S}}
\def\E{{\cal E}}
\def\X{{\cal X}}                        
\def\Y{{\cal Y}}                        
\def\qmbox#1{{\quad\mbox{#1}\quad}}
\def\scp{{\scriptscriptstyle^{\,\circ}}}
\def\sooe{{\textstyle{1\over\eps}}}
\def\FPL{\mbox{FPL} }

\def\leqt{_{1:t}}

\def\leqT{_{1:T}}

\def\ltt{_{<t}}

\def\smin{^{min}}

\def\text#1{\mbox{\scriptsize{#1}}}

\begin{document}

\title{\normalsize\sc Technical Report \hfill IDSIA-08-04
\vskip 2mm\bf\Large\hrule height5pt \vskip 6mm
Prediction with Expert Advice by Following \\
the Perturbed Leader for General Weights
\vskip 6mm \hrule height2pt \vskip 5mm}
\author{{\bf Marcus Hutter} and {\bf Jan Poland}\\[3mm]
\normalsize IDSIA, Galleria 2, CH-6928\ Manno-Lugano, Switzerland%
\thanks{This work was supported by SNF grant 2100-67712.02.}\\
\normalsize \{marcus,jan\}@idsia.ch, \ http://www.idsia.ch/$^{_{_\sim}}\!$\{marcus,jan\} }
\maketitle

\begin{abstract}
When applying aggregating strategies to Prediction with Expert
Advice, the learning rate must be adaptively tuned. The natural
choice of $\sqrt{\mbox{complexity/current loss}}$ renders the
analysis of Weighted Majority derivatives quite complicated. In
particular, for arbitrary weights there have been no results
proven so far. The analysis of the alternative ``Follow the
Perturbed Leader'' (FPL) algorithm from \cite{Kalai:03} (based on
Hannan's algorithm) is easier. We derive loss bounds for adaptive
learning rate and both finite expert classes with uniform weights
and countable expert classes with arbitrary weights. For the
former setup, our loss bounds match the best known results so far,
while for the latter our results are (to our knowledge) new.
\end{abstract}

\begin{keywords}
Prediction with Expert Advice,
Follow the Perturbed Leader,
general weights,
adaptive learning rate,
hierarchy of experts,
expected and high probability bounds,
general alphabet and loss,
online sequential prediction,
\end{keywords}

\newpage
\section{Introduction}\label{secInt}

The theory of Prediction with Expert Advice (PEA)
has rapidly developed in the recent past. Starting with the
Weighted Majority (WM) algorithm of Littlestone and Warmuth
\cite{Littlestone:89,Littlestone:94} and the aggregating strategy
of Vovk \cite{Vovk:90}, a vast variety of different algorithms and
variants have been published. A key parameter in all these
algorithms is the \emph{learning rate}. While this parameter had
to be fixed in the early algorithms such as WM, \cite{Cesa:97}
established the so-called doubling trick to make the learning rate
coarsely adaptive. A little later, incrementally adaptive
algorithms were developed
\cite{Auer:00,Auer:02,Yaroshinsky:04,Gentile:03}. Unfortunately,
the loss bound proofs for the incrementally adaptive WM variants
are quite complex and technical, despite the typically simple and
elegant proofs for a static learning rate.

The complex growing proof techniques also had another consequence:
While for the original WM algorithm, assertions are proven for
countable classes of experts with arbitrary weights, the modern
variants usually restrict to finite classes with uniform weights
(an exception being \cite{Gentile:03}, see the discussion
section). This might be sufficient for many practical purposes but
it prevents the application to more general classes of predictors.
Examples are extrapolating (=predicting) data points with the help
of a polynomial (=expert) of degree $d=1,2,3,...$ --or-- the (from
a computational point of view largest) class of all computable
predictors. Furthermore, most authors have concentrated on
predicting \emph{binary} sequences, often with the 0/1 loss for
$\{0,1\}$-valued and the absolute loss for $[0,1]$-valued
predictions. Arbitrary losses are less common. Nevertheless, it is
easy to abstract completely from the predictions and consider the
resulting losses only. Instead of predicting according to a
``weighted majority'' in each time step, one chooses one
\emph{single} expert with a probability depending on his past
cumulated loss. This is done e.g.\ in \cite{Freund:97}, where an
elegant WM variant, the Hedge algorithm, is analyzed.

A very different, general approach to achieve similar results is
``Follow the Perturbed Leader'' (FPL). The principle dates back to
as early as 1957, now called Hannan's algorithm \cite{Hannan:57}.
In 2003, Kalai and Vempala published a simpler proof of the main
result of Hannan and also succeeded to improve the bound by
modifying the distribution of the perturbation \cite{Kalai:03}.
The resulting algorithm (which they call FPL*) has the same
performance guarantees as the WM-type algorithms for fixed
learning rate, save for a factor of $\sqrt 2$. A major advantage
we will discover in this work is that its analysis remains easy
for an adaptive learning rate, in contrast to the WM derivatives.
Moreover, it generalizes to online decision problems other than
PEA.

In this work, we study the FPL algorithm for PEA. The problems of
WM algorithms mentioned above are addressed: We consider countable
expert classes with arbitrary weights, adaptive learning rate, and
arbitrary losses. Regarding the adaptive learning rate, we obtain
proofs that are simpler and more elegant than for the
corresponding WM algorithms.
(In particular the proof for a self-confident choice of the learning
rate, Theorem \ref{thFPLLDynamic}, is less than half a page).
Further, we prove the first loss bounds for \emph{arbitrary
weights} and adaptive learning rate. Our result even seems to be
the first for \emph{equal weights} and \emph{arbitrary losses},
however the proof technique from \cite{Auer:02} is likely to
carry over to this case.

This paper is structured as follows. In Section~\ref{secSetup} we
give the basic definitions. Sections~\ref{secIFPL} and
\ref{secFFPL} derive the main analysis tools, following the lines
of \cite{Kalai:03}, but with some important extensions. They are
applied in order to prove various upper bounds in
Section~\ref{secBounds}. Section~\ref{secHierarchy} proposes a
hierarchical procedure to improve the bounds for non-uniform
weights. In Section~\ref{secLowFPL}, a lower bound is established.
Section~\ref{secMisc} treats some additional issues. Finally, in
Section~\ref{secConc} we discuss our results, compare them to
references, and state some open problems.

\section{Setup \& Notation}\label{secSetup}

\paradot{Setup}
Prediction with Expert Advice proceeds as follows. We are asked to
perform sequential predictions $y_t\in\Y$ at times $t=1,2,\ldots$.
At each time step $t$, we have access to the predictions
$(y_t^i)_{1\leq i\leq n}$ of $n$ experts $\{e_1,...,e_n\}$. After
having made a prediction, we make some observation $x_t\in\X$, and
a Loss is revealed for our and each expert's prediction.
(E.g.\ the loss might be 1 if the expert made an erroneous
prediction and 0 otherwise. This is the 0/1-loss.) Our goal is to
achieve a total loss ``not much worse" than the best expert, after
$t$ time steps.

We admit $n\in\SetN\cup\{\infty\}$ experts, each of which is
assigned a known complexity $k^i\geq 0$. Usually we require
$\sum_i e^{-k^i}\leq 1$, for instance $k^i=2\ln(i+1)$. Each
complexity defines a weight by means of $e^{-k^i}$ and vice versa.
In the following we will talk rather of complexities than of
weights. If $n$ is finite, then usually one sets $k^i= \ln n$ for
all $i$, this is the case of \emph{uniform complexities/weights}.
If the set of experts is countably infinite ($n=\infty$), uniform
complexities are not possible. The vector of all complexities is
denoted by $k=(k^i)_{1\leq i\leq n}$. At each time $t$, each
expert $i$ suffers a loss\footnote{The setup, analysis and results
easily scale to $s_t^i\in[0,S]$ for $S>0$ other than 1.}
$s_t^i=$Loss$(x_t,y_t^i)\in[0,1]$, and $s_t=(s_t^i)_{1\leq i\leq
n}$ is the vector of all losses at time $t$. Let
$s\ltt=s_1+\ldots+s_{t-1}$ (respectively $s\leqt=s_1+\ldots+s_t$)
be the total past loss vector (including current loss $s_t$) and
$s\leqt\smin=\min_i\{s\leqt^i\}$ be the loss of the \emph{best
expert in hindsight (BEH)}. Usually we do not know in advance the
time $t\geq 0$ at which the performance of our predictions are
evaluated.

\paradot{General decision spaces}
The setup can be generalized as follows. Let $\S\subset\SetR^n$ be the
\emph{state space} and $\D\subset\SetR^n$ the \emph{decision
space}. At time $t$ the state is $s_t\in\S$, and a decision
$d_t\in\D$ (which is made before the state is revealed) incurs a
loss $d_t\!\scp s_t$, where ``$\scp$" denotes the inner product. This
implies that the loss function is \emph{linear} in the states.
Conversely, each linear loss function can be represented in this
way. The decision which minimizes the loss in state $s\in\S$ is
\beq\label{Mdef}
  M(s):=\arg\min_{d\in\D} \{d\scp s\}
\eeq
if the minimum exists. The application of this general framework
to PEA is straightforward: $\D$ is identified with the space of
all unit vectors $\E=\{e_i:1\leq i\leq n\}$, since a decision
consists of selecting a single expert, and $s_t\in[0,1]^n$, so
states are identified with losses. Only Theorem~\ref{thIFPL}
will be stated in terms of general decision space,
where we require that all minima are
attained.\footnote{Apparently, there is no natural condition on
$\D$ and/or $\S$ which guarantees the existence of all minima for
$n=\infty$.} Our main focus is $\D=\E$. However, all our results
generalize to the simplex $\D=\Delta=\{v\in[0,1]^n:\sum_i
v^i=1\}$, since the minimum of a linear function on $\Delta$ is
always attained on $\E$.

\paradot{Follow the Perturbed Leader}
Given $s\ltt$ at time $t$, an immediate idea to solve the expert
problem is to ``Follow the Leader'' (FL), i.e.\ selecting the
expert $e_i$ which performed best in the past (minimizes $s\ltt^i$),
that is predict according to expert $M(s\ltt)$. This approach
fails for two reasons. First, for $n=\infty$ the minimum in
(\ref{Mdef}) may not exist. Second, for $n=2$ and
$s={\,0\,1\,0\,1\,0\,1 \ldots \choose \frac{1}{2}0\,1\,0\,1\,0
\ldots}$, FL always chooses the wrong prediction
\cite{Kalai:03}. We solve the first problem by penalizing each
expert by its complexity, i.e.\ predicting according to expert
$M(s\ltt+k)$. The \emph{FPL (Follow the Perturbed Leader)}
approach solves the second problem by adding to each
expert's loss $s\ltt^i$ a random perturbation.
We choose this perturbation to be negative \emph{exponentially
distributed}, either independent in each time step or once and for
all at the very beginning at time $t=0$. These two possibilities
are equivalent with respect to {\it expected losses}, since the
expectation is linear. The former choice is preferable in order to
protect against an adaptive adversary who generates the $s_t$, and
in order to get bounds with high probability
(Section~\ref{secMisc}). For the main analysis however, the latter is
more convenient. So henceforth we can assume without loss of
generality one initial perturbation $q$ when dealing with expected
loss.

\paranodot{The FPL algorithm} is defined as follows:\\
%
\hspace*{1cm}Choose random vector $q\stackrel{d.}{\sim}\exp$, i.e.\ $P[q^i=u]=e^{-u}$ for all $1\leq i\leq n$. \\
\hspace*{1cm}For $t=1,...,T$\\
\hspace*{1cm}- Choose learning rate $\eps_t$.\\
\hspace*{1cm}- Output prediction of expert $i$ which minimizes $s_{<t}^i+(k^i-q^i)/\eps_t$.\\
\hspace*{1cm}- Receive loss $s_t^i$ for all experts $i$.

\vspace{1.5ex}\noindent Other than $s\ltt$, $k$ and $q$, FPL
depends on the \emph{learning rate} $\eps_t$. We will give choices
for $\eps_t$ in Section~\ref{secBounds}, after having established
the main tools for the analysis. The expected loss at time $t$ of
FPL is
$\ell_t:=E\big[M(s_{<t}+{k-q\over\eps_t})\scp s_t\big]$. The key
idea in the FPL analysis is the use of an intermediate predictor
\emph{IFPL} (for \emph{Implicit or Infeasible FPL}). IFPL predicts
according to $M(s\leqt+{k-q\over\eps_t})$, thus under the
knowledge of $s_t$ (which is of course not available in reality).
By $r_t:=E\big[M(s_{1:t}+{k-q\over\eps_t})\scp s_t\big]$ we
denote the expected loss of IFPL at time $t$. The losses of IFPL
will be upper bounded by BEH in Section~\ref{secIFPL} and lower
bounded by FPL in Section~\ref{secFFPL}.

\paradot{Notes}
Observe that we have stated the FPL algorithm regardless of the
actual \emph{predictions} of the experts and possible
\emph{observations}, only the \emph{losses} are relevant.
Note also that an expert can implement a highly complicated strategy
depending on past outcomes, despite its trivializing
identification with a constant unit vector. The complex expert's
(and environment's) behavior is summarized and hidden in the state
vector $s_t=$Loss$(x_t,y_t^i)_{1\leq i\leq n}$.
Our results therefore apply to \emph{arbitrary prediction and
observation spaces $\Y$ and $\X$ and arbitrary bounded loss
functions}.
This is in contrast to the major part of PEA work
developed for binary alphabet and 0/1 or absolute loss only.
Finally note that the setup allows for losses generated by an
adversary who tries to maximize the regret of FPL and knows the
FPL algorithm and all experts' past predictions/losses. If the
adversary also has access to FPL's past decisions, then FPL must
use independent randomization at each time step in order to
achieve good regret bounds.

\section{IFPL bounded by Best Expert in Hindsight}\label{secExpMax}\label{secIFPL}

In this section we provide tools for comparing the loss of IFPL
to the loss of the best expert in hindsight. The first result
bounds the expected error induced by the exponentially distributed
perturbation.

\begin{lemma}[Maximum of Shifted Exponential Distributions]\label{lemExpMax}
Let $q^1,...,q^n$ be identically exponentially distributed random
variables, i.e.\ $P[q^i]=e^{-q^i}$ for $q^i\geq 0$ and $1\leq
i\leq n\leq\infty$, and $k^i\in\SetR$ be real numbers with
$u:=\sum_{i=1}^n e^{-k^i}$. Then
\bqan
  E[\max_i\{q^i-k^i\}] &\leq& 1+\ln u.
\eqan
\end{lemma}

\paradot{Proof} Using
  $P[q^i\geq b]\leq e^{-b}$ for $b\in\SetR$ we get
\beqn
  P[\max_i\{q^i-k^i\}\geq a]
  = P[\exists i:q^i-k^i\geq a]
  \leq \sum_{i=1}^n P[q^i-k^i\geq a]
  \leq \sum_{i=1}^n e^{-a-k^i} = u\!\cdot\!e^{-a}
\eeqn
where the first inequality is the union bound.
Using $E[z]\leq E[\max\{0,z\}]=\int_0^\infty P[\max\{0,z\}\geq
y]dy = \int_0^\infty P[z\geq y]dy$ (valid for any real-valued
random variable $z$) for $z=\max_i\{q^i-k^i\}-\ln u$, this implies
\beqn
  E[\max_i\{q^i-k^i\}-\ln u]
  \leq \int_0^\infty P\big[\max_i\{q^i-k^i\}\geq y+\ln u \big]dy
  \leq \int_0^\infty e^{-y} dy\ = \ 1,
\eeqn
which proves the assertion. \qed

If $n$ is finite, a lower bound $E[\max_i q^i]\geq 0.57721+\ln n$
can be derived, showing that the upper bound on $E[\max]$ is quite
tight (at least) for $k^i=0$ $\forall i$.
The following bound generalizes \cite[Lem.3]{Kalai:03} to
arbitrary weights.

\begin{theorem}[IFPL bounded by BEH]\label{thIFPL}
Let $\D\subseteq\SetR^n$, $s_t\in\SetR^n$ for $1\leq t\leq T$
(both $\D$ and $s$ may even be negative, but we assume that all
required extrema are attained), and $q,k\in\SetR^n$. If
$\eps_t>0$ is decreasing in $t$, then the loss of the infeasible FPL
knowing $s_t$ at time $t$ in advance (l.h.s.) can be bounded in
terms of the best predictor in hindsight (first term on r.h.s.) plus
additive corrections:
\beqn
  \sum_{t=1}^T M(s_{1:t}+{k\!-\!q\over\eps_t})\scp s_t
  \leq \min_{d\in\D}\{d\scp(s_{1:T}+{k\over\eps_T})\}
     + {1\over\eps_T}\max_{d\in\D}\{d\scp(q-k)\}
     - {1\over\eps_T} M(s_{1:T}+{k\over\eps_T})\scp q.
\eeqn
\end{theorem}

\paradot{Proof} For notational convenience, let $\eps_0=\infty$ and
$\tilde s\leqt=s\leqt+\frac{k-q}{\eps_t}$. Consider the losses
$\tilde s_t=s_t+(k-q)\big(\frac{1}{\eps_t}-\frac{1}{\eps_{t-1}}\big)$
for the moment. We first show by induction on $T$ that the
infeasible predictor $M(\tilde s_{1:t})$ has zero regret, i.e.\
\beq\label{eqnoregret}
  \sum_{t=1}^T M(\tilde s_{1:t})\scp \tilde s_t \leq M(\tilde s_{1:T})\scp \tilde s_{1:T}.
\eeq
For $T=1$ this is obvious. For the induction step from $T-1$ to $T$
we need to show
\beqn
  M(\tilde s_{1:T})\scp \tilde s_T \leq M(\tilde s_{1:T})\scp
  \tilde s_{1:T} - M(\tilde s_{<T})\scp \tilde s_{<T}.
\eeqn
This follows from $\tilde s_{1:T}=\tilde s_{<T}+\tilde s_T$ and
$M(\tilde s_{1:T})\scp \tilde s_{<T} \geq M(\tilde s_{<T})\scp
\tilde s_{<T}$ by minimality of $M$.
Rearranging terms in (\ref{eqnoregret}), we obtain
\beq\label{eqifpl2}
  \sum_{t=1}^T M(\tilde s_{1:t})\scp s_t
  \ \leq\
  M(\tilde s_{1:T})\scp \tilde s_{1:T}- \sum_{t=1}^T M(\tilde s_{1:t})\scp
  (k-q)\Big(\frac{1}{\eps_t}-\frac{1}{\eps_{t-1}}\Big)
\eeq
Moreover, by minimality of $M$,
\bqa
\label{eqifpl4}
M(\tilde s_{1:T})\scp \tilde s_{1:T} & \leq &
M\Big(s_{1:T}+\frac{k}{\eps_T}\Big)\scp
\Big(s_{1:T}+\frac{k-q}{\eps_T}\Big)\\
\nonumber
& = & \min_{d\in\D}\left\{d\scp(s_{1:T}+{k\over\eps_T})\right\}-
M\Big(s_{1:T}+\frac{k}{\eps_T}\Big)\scp
\frac{q}{\eps_T}
\eqa
holds. Using ${1\over\eps_t}-{1\over\eps_{t-1}}\geq 0$ and again
minimality of $M$, we have
\bqa\label{eqifpl3}
\sum_{t=1}^T
({1\over\eps_t}-{1\over\eps_{t-1}})M(\tilde s_{1:t})\scp(q-k) &
\leq & \sum_{t=1}^T
({1\over\eps_t}-{1\over\eps_{t-1}})M(k-q)\scp(q-k)\\
\nonumber
&  = & {1\over\eps_T}M(k-q)\scp(q-k)
= {1\over\eps_T}\max_{d\in\D}\{d\scp(q-k)\}
\eqa
Inserting (\ref{eqifpl4}) and (\ref{eqifpl3}) back into (\ref{eqifpl2})
we obtain the assertion.
\qed

Assuming $q$ random with $E[q^i]=1$ and taking the expectation in
Theorem~\ref{thIFPL}, the last term reduces to
$-{1\over\eps_T}\sum_{i=1}^n M(s_{1:T}+{k\over\eps_T})^i$.
If $\D\geq 0$, the term is negative and may be dropped. In case of
$\D=\E$ or $\Delta$, the last term is identical to
$-{1\over\eps_T}$ (since $\sum_i d^i=1$) and keeping it improves
the bound.
Furthermore, we need to evaluate the expectation of the second to
last term in Theorem~\ref{thIFPL}, namely
$E[\max_{d\in\D}\{d\scp(q-k)\}]$. For $\D=\E$ and $q$ being
exponentially distributed, using Lemma~\ref{lemExpMax}, the
expectation is bounded by $1+\ln u$. We hence get the following
bound:

\begin{corollary}[IFPL bounded by BEH]\label{corIFPL}
For $\D=\E$ and $\sum_i e^{-k^i}\leq 1$ and
$P[q^i]=e^{-q^i}$ for $q\geq 0$ and decreasing $\eps_t>0$, the
expected loss of the infeasible FPL exceeds the loss of expert $i$
by at most $k^i/\eps_T$:
\beqn
  r_{1:T} \;\leq\; s_{1:T}^i + {1\over\eps_T}k^i \qmbox{for all} i\leq n.
\eeqn
\end{corollary}

Theorem~\ref{thIFPL} can be generalized to expert
dependent factorizable $\eps_t\leadsto \eps_t^i=\eps_t\cdot\eps^i$
by scaling $k^i\leadsto k^i/\eps^i$ and $q^i\leadsto q^i/\eps^i$.
Using $E[\max_i\{{q^i-k^i\over\eps^i}\}]\leq
E[\max_i\{q^i-k^i\}]/\min_i\{\eps^i\}$, Corollary~\ref{corIFPL},
generalizes to
\beqn
    E[\sum_{t=1}^T M(s_{1:t}+{k-q\over\eps_t^i})\scp s_t]
    \;\leq\; s_{1:T}^i + {1\over\eps_T^i}k^i + {1\over\eps_T^{min}}
    \quad\forall i,
\eeqn
where $\eps_T^{min}:=\min_i\{\eps_T^i\}$.
For example, for $\eps_t^i=\sqrt{k^i/t}$, additionally assuming
$k^i\geq 1$ $\forall i$, we get the desired bound
$s_{1:T}^i+\sqrt{T\cdot(k^i+1)}$. Unfortunately we were not able
to generalize Theorem~\ref{thFIFPL} to expert-dependent $\eps$,
necessary for the final bound on FPL. In Section~\ref{secHierarchy} we solve this problem by a hierarchy of
experts.

\section{Feasible FPL bounded by Infeasible FPL}\label{secFFPL}

This section establishes the relation between the FPL and IFPL
losses. Recall that $\ell_t=E\big[M(s_{<t}+{k-q\over\eps_t})\scp
s_t\big]$ is the expected loss of FPL at time $t$ and
$r_t=E\big[M(s_{1:t}+{k-q\over\eps_t})\scp s_t\big]$ is the
expected loss of IFPL at time $t$.

\begin{theorem}[FPL bounded by IFPL]\label{thFIFPL}
For $\D=\E$ and $0\leq s_t^i\leq 1$ $\forall i$ and arbitrary
$s_{<t}$ and $P[q^i]=e^{-q^i}$, the expected loss of the feasible FPL is at
most a factor $e^{\eps_t}>1$ larger than for the infeasible FPL:
\beqn
  \ell_t\leq e^{\eps_t}r_t, \qmbox{which implies}
  \ell_{1:T}-r_{1:T}\leq \sum_{t=1}^T\eps_t \ell_t.
\eeqn
Furthermore, if $\eps_t\leq 1$, then also $\ell_t\leq
(1+\eps_t+\eps_t^2)r_t\leq (1+2\eps_t)r_t$.
\end{theorem}

\paradot{Proof}
Let $s=s_{<t}+\sooe k$ be the past cumulative penalized state
vector, $q$ be a vector of exponential distributions, i.e.
$P[q^i]=e^{-q^i}$, and $\eps=\eps_t$.
We now define the random variables $I:=\arg\min_i\{s^i-\sooe q^i\}$ and
$J:=\arg\min_i\{s^i+s_t^i-\sooe q^i\}$, where $0\leq s_t^i\leq 1$
$\forall i$. Furthermore, for fixed vector $x\in\SetR^n$ and fixed
$j$ we define $m:=\min_{i\neq j}\{s^i-\sooe x^i\}\leq \min_{i\neq
j}\{s^i+s_t^i-\sooe x^i\}=:m'$.
With this notation and using the independence of $q^j$ from $q^i$
for all $i\neq j$, we get
\beqn
  P[I=j|q^i=x^i\,\forall i\neq j]
  \;=\; P[s^j-\sooe q^j\leq m|q^i=x^i\,\forall i\neq j]
  \;=\; P[q^j\geq\eps(s^j-m)]
\eeqn
\beqn
  \;\leq\; e^\eps P[q^j\geq\eps(s^j-m+1)]
  \;\leq\; e^\eps P[q^j\geq\eps(s^j+s_t^j-m')]
\eeqn
\beqn
  \;=\; e^\eps P[s^j+s_t^j-\sooe q^j\leq m'|q^i=x^i\,\forall i\neq j]
  \;=\; e^\eps P[J=j|q^i=x^i\,\forall i\neq j],
\eeqn
where we have used $P[q^j\geq a]\leq e^\eps P[q^j\geq a+\eps]$.
Since this bound holds under any condition $x$, it also holds
unconditionally, i.e.\ $P[I=j]\leq e^\eps P[J=j]$. For
$\D=\E$ we have $s_t^I=M(s_{<t}+{k-q\over\eps})\scp s_t$ and
$s_t^J=M(s_{1:t}+{k-q\over\eps})\scp s_t$, which implies
\beqn
  \ell_t
  \;=\;E[s_t^I]
  \;=\; \sum_{j=1}^n s_t^j\!\cdot\!P[I=j]
  \;\leq\; e^\eps \sum_{j=1}^n s_t^j\!\cdot\!P[J=j]
  \;=\; e^\eps E[s_t^J]
  \;=\; e^\eps r_t.
\eeqn
Finally, $\ell_t-r_t\leq\eps_t\ell_t$ follows from $r_t\geq
e^{-\eps_t}\ell_t\geq (1-\eps_t)\ell_t$, and $\ell_t\leq
e^{\eps_t}r_t\leq (1+\eps_t+\eps_t^2)r_t\leq (1+2\eps_t)r_t$ for
$\eps_t\leq 1$ is elementary.
\qed

\paradot{Remark}
As in \cite{Kalai:03}, one can prove a similar statement for
general decision space $\D$ as long as $\sum_i|s_t^i|\leq A$ is
guaranteed for some $A>0$: In this case, we have $\ell_t\leq
e^{\eps_t A}r_t$. If $n$ is finite, then the bound holds for
$A=n$. For $n=\infty$, the assertion holds under the somewhat
unnatural assumption that $\S$ is $l^1$-bounded.

\section{Combination of Bounds and Choices for $\eps_t$}\label{secBounds}

Throughout this section, we assume
\beq
\label{eq:Assumptions}
  \D=\E,\quad s_t\in[0,1]^n\ \forall t,\quad P[q^i]=e^{-q^i}\ \forall i,\ \qmbox{and}
\sum_i e^{-k^i}\leq 1.
\eeq
We distinguish \emph{static} and \emph{dynamic} bounds. Static
bounds refer to a constant $\eps_t\equiv\eps$. Since this value
has to be chosen in advance, a static choice of $\eps_t$ requires
certain prior information and therefore is not practical in many
cases. However, the static bounds are very easy to derive, and
they provide a good means to compare different PEA algorithms. If
on the other hand the algorithm shall be applied without
appropriate prior knowledge, a dynamic choice of $\eps_t$ depending
only on $t$ and/or past observations, is necessary.

\begin{theorem}[FPL bound for static $\eps_t=\eps\propto 1/\sqrt{L}$]\label{thFPLStatic}
Assume (\ref{eq:Assumptions}) holds, then the expected loss
$\ell_t$ of feasible FPL, which employs the prediction of the
expert $i$ minimizing $s_{<t}^i+{k^i-q^i\over\eps_t}$, is bounded
by the loss of the best expert in hindsight in the following way:
\bqan
  i) & &
  \mbox{For}\quad \eps_t=\eps=1/\sqrt{L}
  \qmbox{with} L\geq\ell_{1:T}
  \qmbox{we have}
\\
     & & \nq
  \ell_{1:T}
  \;\leq\; s_{1:T}^i + \sqrt{L}(k^i+1) \quad\forall i
\\
  ii) & & \nq
  \mbox{For}\quad \eps_t=\sqrt{K/L}
  \qmbox{with} L\geq\ell_{1:T}
  \qmbox{and} k^i\leq K \;\forall i
  \qmbox{we have}
\\
    & & \nq
  \ell_{1:T}
  \;\leq\; s_{1:T}^i + 2\sqrt{LK} \quad\forall i
\\
  iii) & & \nq
  \mbox{For}\quad \eps_t=\sqrt{k^i/L}
  \qmbox{with} L\geq \max\{s_{1:T}^i,k^i\}
  \qmbox{we have}
\\
    & & \nq
  \ell_{1:T}
  \;\leq\; s_{1:T}^i + 2\sqrt{Lk^i}+3k^i
\eqan
\end{theorem}

Note that according to assertion $(iii)$, knowledge of only the
\emph{ratio} of the complexity and the loss of the best
expert is sufficient in order to obtain good static bounds, even
for non-uniform complexities.

\paradot{Proof} $(i,ii)$ For $\eps_t=\sqrt{K/L}$ and $L\geq\ell_{1:T}$,
from Theorem~\ref{thFIFPL} and Corollary
\ref{corIFPL}, we get
\beqn
  \ell_{1:T}-r_{1:T}
  \leq \sum_{t=1}^T\eps_t\ell_t
  = \ell_{1:T}\sqrt{K/L}\leq\sqrt{LK}
  \qmbox{and}
  r_{1:T}-s_{1:T}^i
  \leq k^i/\eps_T=k^i\sqrt{L/K}
\eeqn
Combining both, we get
$\ell_{1:T}-s_{1:T}^i\leq\sqrt{L}(\sqrt{K}+k^i/\sqrt{K})$.
$(i)$ follows from $K=1$ and $(ii)$ from $k^i\leq K$.

\noindent
$(iii)$ For $\eps=\sqrt{k^i/L}\leq 1$ we get
\bqan
  \ell_{1:T}
  & \leq & e^\eps r_{1:T}
  \leq (1+\eps+\eps^2)r_{1:T}
  \leq (1+\sqrt{k^i\over L}+{k^i\over L})(s_{1:T}^i+\sqrt{L\over
  k^i}k^i)\\
  & \leq & s_{1:T}^i+\sqrt{Lk^i} +(\sqrt{k^i\over L}+{k^i\over L})(L+\sqrt{Lk^i})
  = s_{1:T}^i + 2\sqrt{Lk^i} +(2+\sqrt{k^i\over L})k^i
\eqan
\qed

The static bounds require knowledge of an upper bound $L$ on the
loss (or the ratio of the complexity of the best expert and its
loss). Since the instantaneous loss is bounded by $1$, one may set
$L=T$ if $T$ is known in advance. For finite $n$ and $k^i=K=\ln
n$, bound $(ii)$ gives the classic regret $\propto\sqrt{T\ln
n}$. If neither $T$ nor $L$ is known, a dynamic choice of $\eps_t$
is necessary. We first present bounds with regret $\propto\sqrt{T}$,
thereafter with regret $\propto\sqrt{s_{1:T}^i}$.

\begin{theorem}[FPL bound for dynamic $\eps_t\propto 1/\sqrt{t}$]\label{thFPLTDynamic}
Assume (\ref{eq:Assumptions}) holds.
\bqan
  i) & & \nq
  \mbox{For}\quad \eps_t=1/\sqrt{t}
  \qmbox{we have}
  \ell_{1:T}
  \;\leq\; s_{1:T}^i + \sqrt{T}(k^i+2) \quad\forall i
\\
  ii) & & \nq
  \mbox{For}\quad \eps_t=\sqrt{K/2t}
  \;\;\mbox{and}\;\; k^i\leq K \;\forall i
  \;\;\mbox{we have}\;\;
  \ell_{1:T}
  \;\leq\; s_{1:T}^i + 2\sqrt{2TK}
  \quad\forall i
\eqan
\end{theorem}

\paradot{Proof} For $\eps_t=\sqrt{K/2t}$, using
$\sum_{t=1}^T{1\over\sqrt{t}}\leq\int_0^T{dt\over\sqrt{t}}=
2\sqrt{T}$ and $\ell_t\leq 1$ we get
\beqn
  \ell_{1:T}-r_{1:T}
  \leq \sum_{t=1}^T \eps_t
  \leq \sqrt{2TK}
  \qmbox{and}
  r_{1:T}-s_{1:T}^i
  \leq {k^i/\eps_T}=k^i\sqrt{2T\over K}
\eeqn
Combining both, we get
$\ell_{1:T}-s_{1:T}^i \leq \sqrt{2T}(\sqrt{K}+k^i/\sqrt{K})$.
$(i)$ follows from $K=2$ and $(ii)$ from $k^i\leq K$.
\qed

In Theorem~\ref{thFPLStatic} we assumed knowledge of an
upper bound $L$ on $\ell_{1:T}$. In an adaptive form,
$L_t:=\ell_{<t}+1$, known at the beginning of time $t$, could be used
as an upper bound on $\ell_{1:t}$ with corresponding adaptive
$\eps_t\propto 1/\sqrt{L_t}$. Such choice of $\eps_t$ is also
called \emph{self-confident} \cite{Auer:02}.

\begin{theorem}[FPL bound for self-confident $\eps_t\propto 1/\sqrt{\ell_{<t}}$]\label{thFPLLDynamic}
Assume (\ref{eq:Assumptions}) holds.
\bqan
  i) & & \nq
  \mbox{For}\quad \eps_t=1/\sqrt{2(\ell_{<t}+1)}
  \qmbox{we have}
\\
   & & \nq
  \ell_{1:T}
  \;\leq\; s_{1:T}^i + (k^i\!+\!1)\sqrt{2(s_{1:T}^i\!+\!1)} + 2(k^i\!+\!1)^2
  \quad\forall i
\\
  ii) & & \nq
  \mbox{For}\quad \eps_t=\sqrt{K/2(\ell_{<t}+1)}
  \qmbox{and} k^i\leq K \;\forall i
  \qmbox{we have}
\\
    & & \nq
  \ell_{1:T}
  \;\leq\; s_{1:T}^i + 2\sqrt{2(s_{1:T}^i\!+\!1)K} + 8K
  \quad\forall i
\eqan
\end{theorem}

\paradot{Proof} Using
$\eps_t=\sqrt{K/2(\ell_{<t}+1)}\leq\sqrt{K/2\ell_{1:t}}$ and
${b-a\over\sqrt
b}=(\sqrt{b}-\sqrt{a})(\sqrt{b}+\sqrt{a}){1\over\sqrt{b}}\leq
2(\sqrt{b}-\sqrt{a})$ for $a\leq b$ and $t_0=\min\{t:\ell_{1:t}>0\}$ we get
\beqn\label{eqLD}
  \ell_{1:T}\!-\!r_{1:T}
  \leq \sum_{t=t_0}^T \eps_t\ell_t
  \leq \sqrt{K\over 2}\sum_{t=t_0}^T {\ell_{1:t}\!-\!\ell_{<t}\over\sqrt{\ell_{1:t}}}
  \leq \sqrt{2K}\sum_{t=t_0}^T [\sqrt{\ell_{1:t\!\!}}\,-\!\sqrt{\ell_{<t\!\!}}\;]
  = \sqrt{2K}\sqrt{\ell_{1:T}}
\eeqn
Adding
$r_{1:T}-s_{1:T}^i \leq {k^i\over\eps_T} \leq
k^i\sqrt{2(\ell_{1:T}+1)/K}$ we get
\beqn
  \ell_{1:T}-s_{1:T}^i
  \leq \sqrt{2\bar\kappa^i(\ell_{1:T}\!+\!1)},
  \qmbox{where}
  \sqrt{\bar\kappa^i}:=\sqrt{K}+k^i/\sqrt{K}.
\eeqn
Taking the square and solving the resulting quadratic inequality
w.r.t.\ $\ell_{1:T}$ we get
\beqn
  \ell_{1:T}
  \leq s_{1:T}^i + \bar\kappa^i + \sqrt{2(s_{1:T}^i\!+\!1)\bar\kappa^i+(\bar\kappa^i)^2}
  \leq s_{1:T}^i + \sqrt{2(s_{1:T}^i\!+\!1)\bar\kappa^i} + 2\bar\kappa^i
\eeqn
For $K=1$ we get $\sqrt{\bar\kappa^i}=k^i+1$ which yields $(i)$.
For $k^i\leq K$ we get $\bar\kappa^i\leq 4K$ which yields $(ii)$.
\qed

The proofs of results similar to $(ii)$ for WM
for 0/1 loss all fill several pages \cite{Auer:02,Yaroshinsky:04}.
The next result establishes a similar bound, but instead of using
the \emph{expected} value $\ell\ltt$, the \emph{best loss so far}
$s\ltt\smin$ is used. This may have computational advantages,
since $s\ltt\smin$ is immediately available, while $\ell\ltt$
needs to be evaluated (see discussion in Section~\ref{secMisc}).

\begin{theorem}[FPL bound for adaptive $\eps_t\propto 1/\sqrt{s\ltt\smin}$]\label{thFPL2}
Assume (\ref{eq:Assumptions}) holds.
\bqan
  i) & & \nq
  \mbox{For}\quad \eps_t = 1/\min_i\{k^i+\sqrt{(k^i)^2+2s^i\ltt+2}\}
  \qmbox{we have}
\\
   & & \nq
  \ell\leqT \;\leq\; s\leqT^i+(k^i\!+2)\sqrt{2s\leqT^i}+2(k^i\!+2)^2
  \quad \forall i
\\
  ii) & & \nq
  \mbox{For}\quad \eps_t =
  \sqrt{\odt}\!\cdot\!\min\{1,\sqrt{K/s\ltt\smin}\}
  \qmbox{and} k^i\leq K \;\forall i
  \qmbox{we have}
\\
    & & \nq
  \ell\leqT \;\leq\;
  s\leqT\smin+2\sqrt{2K s\leqT\smin}+5K\ln(s\leqT\smin)+3K+6.
\eqan
\end{theorem}
We briefly motivate the strangely looking choice for $\eps_t$ in
$(i)$. The first naive candidate, $\eps_t\propto 1/\sqrt{s\ltt^{min}}$,
turns out too large. The next natural trial is requesting
$\eps_t=1/\sqrt{2\min\{s\ltt^i+\frac{k^i}{\eps_t}\}}$. Solving
this equation results in $\eps_t=1/(k^i+\sqrt{(k^i)^2+2s\ltt^i})$,
where $i$ be the index for which $s\ltt^i+\frac{k^i}{\eps_t}$ is
minimal.

\paradot{Proof} Similar to the proof of the previous theorem,
but more technical.\qed

The bound $(i)$ is a complete square, and also the bounds of
Theorem~\ref{thFPLLDynamic} when adding 1 to them. Hence the
bounds can be written as
$\sqrt{\ell_{1:T}}\leq\sqrt{s_{1:T}^i}+\sqrt{2}(k^i+2)$ and
$\sqrt{\ell_{1:T}}\leq\sqrt{s_{1:T}^i+1}+\sqrt{8K}$ and
$\sqrt{\ell_{1:T}}\leq\sqrt{s_{1:T}^i+1}+\sqrt{2}(k^i+1)$,
respectively, hence the $\sqrt{\mbox{Loss}}$-regrets are bounded
for $T\to\infty$.

\paradot{Remark}
The same analysis as for Theorems
[\ref{thFPLStatic}-\ref{thFPL2}]$(ii)$ applies to general $\D$,
using $\ell_t\leq e^{\eps_t n}r_t$ instead of $\ell_t\leq
e^{\eps_t}r_t$, and leading to an additional factor $\sqrt{n}$ in
the regret. Compare the remark at the end of Section~\ref{secFFPL}.

\section{Hierarchy of Experts}\label{secHierarchy}

We derived bounds which do not need prior knowledge of $L$ with
regret $\propto\sqrt{TK}$ and $\propto\sqrt{s_{1:T}^i K}$ for a
finite number of experts with equal penalty $K=k^i=\ln n$. For
an infinite number of experts, unbounded expert-dependent complexity
penalties $k^i$ are necessary (due to constraint $\sum_i
e^{-k^i}\leq 1$). Bounds for this case (without prior knowledge of
$T$) with regret $\propto k^i\sqrt{T}$ and $\propto
k^i\sqrt{s_{1:T}^i}$ have been derived. In this case, the
complexity $k^i$ is no longer under the square root. It is likely that
improved regret bounds $\propto\sqrt{Tk^i}$ and
$\propto\sqrt{s_{1:T}^i k^i}$ as in the finite case hold. We were
not able to derive such improved bounds for FPL, but for a
(slight) modification. We consider a two-level hierarchy of experts.
First consider an FPL for the subclass of experts of complexity
$K$, for each $K\in\SetN$. Regard these FPL$^K$ as (meta) experts
and use them to form a (meta) FPL. The class of meta experts now
contains for each complexity only one (meta) expert, which allows
us to derive good bounds. In the following, quantities referring
to complexity class $K$ are superscripted by $K$, and meta
quantities are superscripted by $\;\widetilde{}$ .

Consider the class of experts $\E^K:=\{i:K-1<k^i\leq K\}$ of
complexity $K$, for each $K\in\SetN$. FPL$^K$ makes randomized
prediction
$I_t^K:=\arg\min_{i\in\E^K}\{s_{<t}^i+{k^i-q^i\over\eps_t^K}\}$
with $\eps_t^K:=\sqrt{K/2t}$ and suffers loss $u_t^K:=s_t^{I_t^K}$
at time $t$. Since $k^i\leq K$ $\forall i\in\E^k$ we can apply
Theorem~\ref{thFPLTDynamic}$(ii)$ to FPL$^K$:
\beq\label{eqFH}
  E[u_{1:T}^K] \;=\; \ell_{1:T}^K \;\leq\; s_{1:T}^i+ 2\sqrt{2TK}
  \quad \forall i\in\E^K
  \quad \forall K\in\SetN
\eeq
We now define a meta state $\tilde s_t^K=u_t^K$ and regard FPL$^K$
for $K\in\SetN$ as meta experts, so meta expert $K$ suffers loss
$\tilde s_t^K$. (Assigning expected loss $\tilde
s_t^K=E[u_t^K]=\ell_t^K$ to FPL$^K$ would also work.) Hence the
setting is again an expert setting and we define the meta
$\widetilde{\mbox{FPL}}$ to predict $\tilde
I_t:=\arg\min_{K\in\SetN}\{\tilde s_{<t}^K+{\tilde k^K-\tilde
q^K\over \tilde\eps_t}\}$ with $\tilde\eps_t=1/\sqrt{t}$ and
$\tilde k^K=\odt+2\ln K$ (implying $\sum_{K=1}^\infty e^{-\tilde
k^K}\leq 1$). Note that $\tilde s_{1:t}^K=\tilde s_1^K+...+\tilde
s_t^K= s_1^{I_1^K}+...+s_t^{I_t^K}$ sums over the same meta state
components $K$, but over different components ${I_t^K}$ in normal
state representation.

By Theorem~\ref{thFPLTDynamic}$(i)$ the $\tilde q$-expected loss
of $\widetilde{\mbox{FPL}}$ is bounded by $\tilde s_{1:T}^K +
\sqrt{T}(\tilde k^K+2)$. As this bound holds for all $q$ it also holds
in $q$-expectation. So if we define $\tilde\ell_{1:T}$ to be the
$q$ {\em and} $\tilde q$ expected loss of
$\widetilde{\mbox{FPL}}$, and chain this bound with (\ref{eqFH})
for $i\in\E^K$ we get:
\bqan
  \tilde\ell_{1:T}
  &\leq& E[\tilde s_{1:T}^K + \sqrt{T}(\tilde k^K\!+2)]
   \;=\; \ell_{1:T}^K + \sqrt{T}(\tilde k^K\!+2) \\
  &\leq& s_{1:T}^i+ \sqrt{T}[2\sqrt{2(k^i\!+1)}+\odt+2\ln (k^i\!+1)+2],
\eqan
where we have used $K\leq k^i+1$. This bound is valid for all $i$
and has the desired regret $\propto\sqrt{T k^i}$. Similarly we can
derive regret bounds $\propto\sqrt{s_{1:T}^i k^i}$ by exploiting
that the bounds are concave in $s_{1:T}^i$ and using Jensen's
inequality.

\begin{theorem}[Hierarchical FPL bound for dynamic $\eps_t$]\label{thHFPL}
The hierarchical $\widetilde{\FPL}$ employs at time $t$
the prediction of expert $i_t:=I_t^{\tilde I_t}$, where
\vspace{-0.5ex}\beqn
  I_t^K:=\mathop{\arg\min}_{i:\lceil k^i\rceil=K}\{s_{<t}^i+{\textstyle{k^i-q^i\over\eps_t^K}}\}
  \qmbox{and}
  \tilde I_t:=\mathop{\arg\min}_{K\in\SetN}\Big\{s_1^{I_1^K}+...+s_{t-1}^{I_{t-1}^K}+
  {\textstyle{{1\over 2}+2\ln K -\tilde q^K\over \tilde\eps_t}}\Big\}
  \vspace{-1.5ex}
\eeqn
Under assumptions (\ref{eq:Assumptions}) and $P[\tilde q^K]=e^{-\tilde
q^K}$ $\forall K\in\SetN$, the
expected loss $\tilde\ell_{1:T}=E[s_1^{i_1}+...+s_T^{i_T}]$ of
$\widetilde{\FPL}$ is bounded as follows:
\bqan
  a) & & \nq
  \mbox{For}\quad \eps_t^K=\sqrt{K/2t}
  \qmbox{and} \tilde\eps_t=1/\sqrt{t}
  \qmbox{we have}
\\
   & & \nq
  \tilde\ell_{1:T}
  \;\leq\; s_{1:T}^i + 2\sqrt{2Tk^i}\!\cdot\!\big(1+O({\textstyle{\ln k^i\over \sqrt{k^i}}})\big)
  \quad\forall i.
\\
  b) & & \nq
  \mbox{For $\tilde\eps_t$ as in $(i)$ and $\eps_t^K$ as in $(ii)$
  of Theorem $\{{\ref{thFPLLDynamic}\atop\ref{thFPL2}}\}$ we have}
\\
    & & \nq
  \tilde\ell_{1:T}
  \;\leq\; s_{1:T}^i + 2\sqrt{2s_{1:T}^i k^i}\!\cdot\!\big(1+O({\textstyle{\ln k^i\over \sqrt{k^i}}})\big)
  + {\textstyle\big\{{O(k^i)\atop O(k^i\ln s_{1:T}^i)}\big\}}
  \quad\forall i.
\eqan
\end{theorem}

The hierarchical $\widetilde{\FPL}$ differs from a
direct \FPL over all experts $\E$. One potential way to prove a
bound on direct \FPL may be to show (if it holds) that \FPL
performs better than $\widetilde{\FPL}$, i.e.\ $\ell_{1:T}\leq
\tilde\ell_{1:T}$. Another way may be to suitably generalize
Theorem~\ref{thFIFPL} to expert dependent $\eps$.

\section{Miscellaneous}\label{secMisc}\label{secComp}

\paradot{Lower Bound on FPL}\label{secLowFPL}
For finite $n$, a lower bound on FPL similar to the upper bound in Theorem
\ref{thIFPL} can also be proven.
For any $\D\subseteq\SetR^n$ and $s_t\in\SetR$
such that the required extrema exist, $q\in\SetR^n$, and $\eps_t>0$
decreasing, the loss of FPL for uniform
complexities can be lower bounded in terms of the best
predictor in hindsight plus/minus additive
corrections:
\beq\label{thLowFPL}
  \sum_{t=1}^T M(s_{<t}-{q\over\eps_t})\scp s_t
  \geq \min_{d\in\D}\{d\scp s_{1:T}\}
     - {1\over\eps_T}\max_{d\in\D}\{d\scp q\}
     + \sum_{t=1}^T ({1\over\eps_t}\!-\!{1\over\eps_{t-1}}) M(s_{<t})\scp q
\eeq
For $\D=\E$ and any $\S$ and all $k^i$ equal and $P[q^i]=e^{-q^i}$
for $q\geq 0$ and decreasing $\eps_t>0$, this reduces to
\beq\label{corLowFPL}
  \ell_{1:t} \;\geq\; s_{1:T}^{min} - {\ln n\over\eps_T}
\eeq
The upper and lower bounds on $\ell_{1:T}$
(Theorem~\ref{thFIFPL} and Corollary~\ref{corIFPL} and
(\ref{corLowFPL})) together show that
\beq\label{eqltos}
  {\ell_{1:t}\over s_{1:t}^{min}} \to 1
  \quad\qmbox{if}\quad
  \eps_t\to 0
  \qmbox{and}
  \eps_t\!\cdot\!s_{1:t}^{min} \to \infty
  \qmbox{and}
  k^i=K\;\forall i.
\eeq
For instance, $\eps_t=\sqrt{K/2 s_{<t}^{min}}$. For
$\eps_t=\sqrt{K/2(\ell_{<t}+1)}$ we proved the bound in Theorem
\ref{thFPLLDynamic}$(ii)$. Knowing that $\sqrt{K/2(\ell_{<t}+1)}$
converges to $\sqrt{K/2 s_{<t}^{min}}$ due to (\ref{eqltos}), we
can derive a bound similar to Theorem~\ref{thFPLLDynamic}$(ii)$
for $\eps_t=\sqrt{K/2 s_{<t}^{min}}$. This choice for $\eps_t$ has
the advantage that we do not have to compute $\ell_{<t}$ (see
below), as also achieved by Theorem~\ref{thFPL2}$(ii)$.
We do not know whether (\ref{thLowFPL}) can be
generalized to expert dependent complexities $k^i$.

\paradot{Initial versus independent randomization}
So far we assumed that the perturbations are sampled only once
at time $t=0$. As already indicated, under the expectation this is
equivalent to generating a new perturbation $q_t$ at each time
step $t$. While the former way is favorable for the analysis, the
latter may have two advantages. First, if the losses are generated
by an adaptive adversary, then he may after some time figure out
the random perturbation and use it to force FPL to have a large
loss. Second, repeated sampling of the perturbations guarantees
better bounds with high probability.

\paradot{Bounds with high probability}
We have derived several bounds for the expected loss $\ell_{1:T}$
of FPL. The {\em actual} loss at time $t$ is
$u_t=M(s_{<t}+{k-q\over\eps_t})\scp s_t$. A simple Markov inequality shows
that the total actual loss $u_{1:T}$ exceeds
the total expected loss $\ell_{1:T}=E[u_{1:T}]$ by a factor of
$c>1$ with probability at most $1/c$:
\beqn
  P[u_{1:T}\geq c\!\cdot\!\ell_{1:T}]
  \;\leq\; {1/c}
\eeqn
Randomizing independently for each $t$ as described in the
previous paragraph, the actual loss is
$u_t=M(s_{<t}+{k-q_t\over\eps_t})\scp s_t$ with the same expected loss
$\ell_{1:T}=E[u_{1:T}]$ as before. The advantage of independent
randomization is that we can get a much better high-probability
bound. We can exploit a Chernoff-Hoeffding bound
\cite[Cor.5.2b]{McDiarmid:89}, valid for arbitrary independent
random variables $0\leq u_t\leq 1$ for $t=1,...,T$:
\beqn
  P\Big[|u_{1:T}-E[u_{1:T}]|\geq\delta E[u_{1:T}]\Big]
  \;\leq\; 2\exp(-{\textstyle{1\over 3}}\delta^2 E[u_{1:T}]), \qquad 0\leq\delta\leq 1.
\eeqn
For $\delta=\sqrt{3c/\ell_{1:T}}$ we get
\beq\label{eqCH}
  P[|u_{1:T}-\ell_{1:T}|\geq\sqrt{3c\ell_{1:T}}]
  \;\leq\; 2e^{-c}
  \qmbox{as soon as}
  \ell_{1:T}\geq 3c.
\eeq
Using (\ref{eqCH}), the bounds for $\ell_{1:T}$ of Theorems
\ref{thFPLStatic}-\ref{thFPL2} can be rewritten to yield
similar bounds with high probability ($1-2e^{-c}$) for $u_{1:T}$
with small extra regret $\propto\sqrt{c\cdot L}$ or $\propto\sqrt{c\cdot
s_{1:T}^i}$.
Furthermore, (\ref{eqCH}) shows that with high probability,
$u_{1:T}/\ell_{1:T}$ converges rapidly to 1 for
$\ell_{1:T}\to\infty$. Hence we may use the easier to compute
$\eps_t=\sqrt{K/2u_{<t}}$ instead of
$\eps_t=\sqrt{K/2(\ell_{<t}+1)}$, with similar bounds on the
regret.

\paradot{Computational Aspects}
It is easy to generate the randomized decision of FPL. Indeed,
only a single initial exponentially distributed vector
$q\in\SetR^n$ is needed. Only for adaptive $\eps_t\propto
1/\sqrt{\ell_{<t}}$ (see Theorem~\ref{thFPLLDynamic}) we need to
compute expectations explicitly. Given $\eps_t$, from $t\leadsto
t+1$ we need to compute $\ell_t$ in order to update $\eps_t$. Note
that $\ell_t=w_t\!\scp s_t$, where $w_t^i=P[I_t=i]$ and
$I_t:=\arg\min_{i\in\E}\{s_{<t}^i+{k^i-q^i\over\eps_t}\}$ is the
actual (randomized) prediction of FPL. With $s:=s_{<t}+k/\eps_t$,
$P[I_t=i]$ has the following representation:
\beqn
  P[I_t\!=i]
  \;= \int_{-\infty}^{s^{min}} \!\!\nq\eps_t e^{-\eps_t(s^i-m)}
      \prod_{j\neq i}(1-e^{-\eps_t(s^j\!-\!m)})dm \\[-1ex]
  \;=\; \nq\nq\sum_{{\cal M}:\{i\}\subseteq{\cal M}\subseteq{\cal N}}\nq\nq\;
  {\textstyle{(-)^{|{\cal M}|-1}\over|{\cal M}|}}e^{-\eps_t\!\!\sum_{j\in\cal M}(s^j\!-\!s^{min})}
\eeqn
In the last equality we expanded the product and performed the
resulting exponential integrals. For finite $n$, the
one-dimensional integral should be numerically feasible. Once the
product $\prod_{j=1}^n(1-e^{-\eps_t(s^j-m)})$ has been computed in
time $O(n)$, the argument of the integral can be computed for each
$i$ in time $O(1)$, hence the overall time to compute $\ell_t$ is
$O(c\cdot n)$, where $c$ is the time to numerically compute one
integral. For infinite%
\footnote{
For practical realizations in case of infinite $n$, one must use
finite subclasses of increasing size, compare
\cite{Littlestone:94}. }
$n$, the last sum may be approximated by the dominant
contributions. The expectation may also be approximated by (monte
carlo) sampling $I_t$ several times.
Recall that approximating $\ell\ltt$ can be avoided by
using $s\ltt\smin$ (Theorem~\ref{thFPL2}) or $u\ltt$ (bounds with
high probability) instead.

\paradot{Deterministic prediction and absolute loss}
Another use of $w_t$ from the last paragraph is the following: If
the decision space is $\D=\Delta$, then FPL may make a
deterministic decision $d=w_t\in\Delta$ at time $t$ with bounds
now holding for sure, instead of selecting $e_i$ with probability
$w_t^i$. For example for the absolute loss $s_t^i=|x_t-y_t^i|$
with observation $x_t\in[0,1]$ and predictions $y_t^i\in[0,1]$, a
master algorithm predicting deterministically $w_t\!\scp
y_t\in[0,1]$ suffers absolute loss $|x_t-w_t\!\scp y_t|\leq\sum_i
w_t^i|x_t-y_t^i|=\ell_t$, and hence has the same (or better)
performance guarantees as FPL. In general, masters can be chosen
deterministic if prediction space $\Y$ and loss-function Loss$(x,y)$ are
convex.

\section{Discussion and Open Problems}\label{secConc}

How does FPL compare with other expert advice algorithms? We
briefly discuss four issues.

\paradot{Static bounds}
Here the coefficient of the regret term $\sqrt{KL}$, referred to
as the \emph{leading constant} in the sequel, is $2$ for FPL
(Theorem~\ref{thFPLStatic}). It is thus a factor of $\sqrt 2$
worse than the Hedge bound for arbitrary loss \cite{Freund:97},
which is sharp in some sense \cite{Vovk:95}. For special loss
functions, the bounds can sometimes be improved, e.g. to a leading
constant of 1 in the static WM case with 0/1 loss \cite{Cesa:97}.

\paradot{Dynamic bounds}
Not knowing the right learning rate in advance usually costs a
factor of $\sqrt 2$. This is true for Hannan's algorithm
\cite{Kalai:03} as well as in all our cases. Also for binary
prediction with uniform complexities and 0/1 loss, this result has
been established recently -- \cite{Yaroshinsky:04} show a dynamic
regret bound with leading constant $\sqrt 2(1+\eps)$. Remarkably,
the best dynamic bound for a WM variant proven in \cite{Auer:02}
has a leading constant $2\sqrt 2$, which matches ours. Considering
the difference in the static case, we therefore conjecture that a
bound with leading constant of $2$ holds for a dynamic Hedge
algorithm.

\paradot{General weights}
While there are several dynamic bounds for uniform weights, the
only result for non-uniform weights we know of is
\cite[Cor.16]{Gentile:03}, which gives a dynamic bound for a
$p$-norm algorithm for the absolute loss if the weights are
rapidly decaying. Our hierarchical FPL bound in Theorem
\ref{thHFPL} $(b)$ generalizes it to arbitrary weights and losses
and strengthens it, since both, asymptotic order and leading
constant, are smaller. Also the FPL analysis gets more complicated
for general weights. We conjecture that the bounds
$\propto\sqrt{Tk^i}$ and $\propto\sqrt{s_{1:T}^i k^i}$ also hold
without the hierarchy trick, probably by using expert dependent
learning rate $\eps_t^i$.

\paradot{Comparison to Bayesian sequence prediction}
We can also compare the \emph{worst-case} bounds for FPL obtained
in this work to similar bounds for \emph{Bayesian sequence
prediction}. Let $\{\nu_i\}$ be a class of probability
distributions over sequences and assume that the true sequence is
sampled from $\mu\in\{\nu_i\}$ with complexity $k^\mu$ ($\sum_i
2^{-k^{\nu_i}}\leq 1$). Then it is known that the Bayes-optimal
predictor based on the $2^{-k^{\nu_i}}$-weighted mixture of
$\nu_i$'s has an expected total loss of at most
$L^\mu+2\sqrt{L^\mu k^\mu}+2k^\mu$, where $L^\mu$ is the expected
total loss of the Bayes-optimal predictor based on $\mu$
\cite[Thm.2]{Hutter:02spupper}. Using FPL, we obtained the same
bound except for the leading order constant, but for any sequence
independently of the assumption that it is generated by $\mu$.
This is another indication that a PEA bound with leading constant
2 could hold. See \cite[Sec.6.3]{Hutter:03optisp} for a more
detailed comparison of Bayes bounds with PEA bounds.


\begin{small}

\end{small}


\begin{thebibliography}{ACBG02}

\bibitem[ACBG02]{Auer:02}
P.~Auer, N.~Cesa-Bianchi, and C.~Gentile.
\newblock Adaptive and self-confident on-line learning algorithms.
\newblock {\em Journal of Computer and System Sciences}, 64(1):48--75, 2002.

\bibitem[AG00]{Auer:00}
P.~Auer and C.~Gentile.
\newblock Adaptive and self-confident on-line learning algorithms.
\newblock In {\em Proceedings of the 13th Conference on Computational Learning
  Theory}, pages 107--117. Morgan Kaufmann, San Francisco, 2000.

\bibitem[CB97]{Cesa:97}
N.~Cesa-Bianchi{ et al.}
\newblock How to use expert advice.
\newblock {\em Journal of the ACM}, 44(3):427--485, 1997.

\bibitem[FS97]{Freund:97}
Y.~Freund and R.~E. Schapire.
\newblock A decision-theoretic generalization of on-line learning and an
  application to boosting.
\newblock {\em Journal of Computer and System Sciences}, 55(1):119--139, 1997.

\bibitem[Gen03]{Gentile:03}
C.~Gentile.
\newblock The robustness of the p-norm algorithm.
\newblock {\em Machine Learning}, 53(3):265--299, 2003.

\bibitem[Han57]{Hannan:57}
J.~Hannan.
\newblock Approximation to {Bayes} risk in repeated plays.
\newblock In M.~Dresher, A.~W. Tucker, and P.~Wolfe, editors, {\em
  Contributions to the Theory of Games 3}, pages 97--139. Princeton University
  Press, 1957.

\bibitem[Hut03a]{Hutter:02spupper}
M.~Hutter.
\newblock Convergence and loss bounds for {Bayesian} sequence prediction.
\newblock {\em IEEE Transactions on Information Theory}, 49(8):2061--2067,
  2003.

\bibitem[Hut03b]{Hutter:03optisp}
M.~Hutter.
\newblock Optimality of universal {B}ayesian prediction for general loss and
  alphabet.
\newblock {\em Journal of Machine Learning Research}, 4:971--1000, 2003.

\bibitem[KV03]{Kalai:03}
A.~Kalai and S.~Vempala.
\newblock Efficient algorithms for online decision.
\newblock In {\em Proceedings of the 16th Annual Conference on Learning Theory
  ({COLT-2003})}, Lecture Notes in Artificial Intelligence, pages 506--521,
  Berlin, 2003. Springer.

\bibitem[LW89]{Littlestone:89}
N.~Littlestone and M.~K. Warmuth.
\newblock The weighted majority algorithm.
\newblock In {\em 30th Annual Symposium on Foundations of Computer Science},
  pages 256--261, Research Triangle Park, North Carolina, 1989. IEEE.

\bibitem[LW94]{Littlestone:94}
N.~Littlestone and M.~K. Warmuth.
\newblock The weighted majority algorithm.
\newblock {\em Information and Computation}, 108(2):212--261, 1994.

\bibitem[McD89]{McDiarmid:89}
C.~McDiarmid.
\newblock On the method of bounded differences.
\newblock {\em Surveys in Combinatorics}, 141, London Mathematical Society
  Lecture Notes Series:148--188, 1989.

\bibitem[Vov90]{Vovk:90}
V.~G. Vovk.
\newblock Aggregating strategies.
\newblock In {\em Proceedings of the Third Annual Workshop on Computational
  Learning Theory}, pages 371--383, Rochester, New York, 1990. ACM Press.

\bibitem[Vov95]{Vovk:95}
V.~G. Vovk.
\newblock A game of prediction with expert advice.
\newblock In {\em Proceedings of the 8th Annual Conference on Computational
  Learning Theory}, pages 51--60. ACM Press, New York, NY, 1995.

\bibitem[YEYS04]{Yaroshinsky:04}
R.~Yaroshinsky, R.~El-Yaniv, and S.~Seiden.
\newblock How to better use expert advice.
\newblock {\em Machine Learning}, 2004.

\end{thebibliography}
\end{document}